%% file: CRIHP.tex
\documentclass[conference]{IEEEtran}
\IEEEoverridecommandlockouts
\usepackage{cite}
\usepackage{algpseudocode} 
\usepackage{balance}
\usepackage{soul}
\usepackage{array}
\usepackage[nointegrals]{wasysym}
\usepackage{tikz}
\usepackage{times}
\usepackage{soul}
\usepackage{url}
\usepackage[hidelinks]{hyperref}
\usepackage[utf8]{inputenc}
\usepackage[small]{caption}
\usepackage{graphicx}
\usepackage{booktabs}
\usepackage{dsfont}
\usepackage{amssymb}
\usepackage{amsmath}
\usepackage{multirow}
\usepackage{enumitem}
\usepackage{upgreek}
\usepackage[ruled,linesnumbered]{algorithm2e}
\usepackage{xcolor}
\usepackage{soul}
\usepackage[font=small,skip=5pt]{caption}
\usepackage[font=small,skip=5pt,labelformat=simple]{subcaption}
\usepackage{float}
\usepackage{diagbox}
\usepackage{pbox}
\usepackage{dblfloatfix}

\usepackage{amsmath}
\usepackage{hhline}

\def\BibTeX{{\rm B\kern-.05em{\sc i\kern-.025em b}\kern-.08em
    T\kern-.1667em\lower.7ex\hbox{E}\kern-.125emX}}

\begin{document}

\title{Enhancing Asynchronous Time Series Forecasting with Contrastive Relational Inference}

\author{
    \IEEEauthorblockN{Yan Wang\textsuperscript{\dag}, 
      Zhixuan Chu\textsuperscript{\dag},
      Tao Zhou, \thanks{\textsuperscript{\dag}These authors contributed equally.}
      Caigao Jiang,
      Hongyan Hao,
      Minjie Zhu,
      Xindong Cai, \\
      Qing Cui,
      Longfei Li,
      James Y Zhang,
      Siqiao Xue\textsuperscript{\ddag},
      Jun Zhou \thanks{\textsuperscript{\ddag}Corresponding author.}}
    \IEEEauthorblockA{Ant Group \\
    \{luli.wy, chuzhixuan.czx, zt339591\}@antgroup.com
    }
}

\maketitle
\thispagestyle{plain}
\pagestyle{plain}

\begin{abstract}

Asynchronous time series, also known as temporal event sequences, are the basis of many applications throughout different industries. Temporal point processes(TPPs) are the standard method for modeling such data. Existing TPP models have focused on parameterizing the conditional distribution of future events instead of explicitly modeling event interactions, imposing challenges for event predictions. In this paper, we propose a novel approach that leverages Neural Relational Inference (NRI) to learn a relation graph that infers interactions while simultaneously learning the dynamics patterns from observational data. Our approach, the Contrastive Relational Inference-based Hawkes Process (CRIHP), reasons about event interactions under a variational inference framework. It utilizes intensity-based learning to search for prototype paths to contrast relationship constraints. Extensive experiments on three real-world datasets demonstrate the effectiveness of our model in capturing event interactions for event sequence modeling tasks. Code will be integrated into the \href{https://github.com/ant-research/EasyTemporalPointProcess}{\textcolor{blue}{EasyTPP}}\footnote[1]{\url{https://github.com/ant-research/EasyTemporalPointProcess}.} framework.

\end{abstract}

\input{chapters_crihp/_introduction}
\input{chapters_crihp/_relatedwork.tex}
\input{chapters_crihp/_method.tex}
\input{chapters_crihp/_experiment.tex}

\section{Conclusion}
This paper introduces CRIHP, a novel framework for relational inference in asynchronous time series forecasting. CRIHP incorporates a contrastive relational inference architecture to model interactions between events without needing additional information. Experiments in real-world settings demonstrate that CRIHP achieves superior performance for this task.

\bibliographystyle{IEEEtran}
\bibliography{references}

\end{document}

%% file: chapters_crihp/_introduction.tex
\section{Introduction}
\label{sec:intro}
Asynchronous time series, also named temporal event sequence data in some literature,  is ubiquitous in daily life, containing discrete events with varying marks and irregular inter-event time intervals. In this work, we focus on the task of event sequence forecasting, which leverages historical sequences to uncover interactions between successive events and predict future events' marks and arrival times. Recently, neural TPPs \cite{12:conf/nips/MeiE17, 50:conf/nips/XueSZM22} begin to show advantages in modeling event sequence, but existing methods mainly focus on parametrizing the conditional distribution of the next event \cite{17:conf/ijcai/ShchurTJG21, 49:journals/corr/abs-2307-08097}, and do not fully discuss how to model the relational structure between events. RNN-based TPPs \cite{11:conf/kdd/DuDTUGS16, 12:conf/nips/MeiE17, 13:journals/tnn/XiaoYFSYZ19} achieve significant progress in event forecasting task, but related methods ignore the explicit modeling of the relationship between events, which is difficult to help us intuitively understand the interactions between events. The Attention-based TPPs \cite{14:zhang2020self, 15:conf/icml/ZuoJLZZ20} use the matching function to construct the similarity coefficient between events, which is not a direct description of the interactions between events. Recently, some work has begun to focus on the relational Inference between events, methods based on Causal Inference \cite{18:conf/icml/XuFZ16, 19:conf/icml/AchabBGMM17, 20:eichler2017graphical,chu2023causal,chu2020matching} define the interaction between events by establishing granger causality, but such models usually make strong assumptions and are difficult to learn different types of influence relationships. Methods based on Graph Neural Networks (GNN) \cite{21:wu2020modeling, 22:conf/aaai/ShangS19, 23:conf/ijcnn/XueSHMZWW21, 24:conf/kdd/LiLKP21} typically construct static graphs according to the mark information. However, in real-world systems, the interactions between events evolve dynamically over time. Existing GNN-based methods struggle to model such dynamic changes in relationships. 

\begin{figure}[h]
\centering
\includegraphics[width=\linewidth]{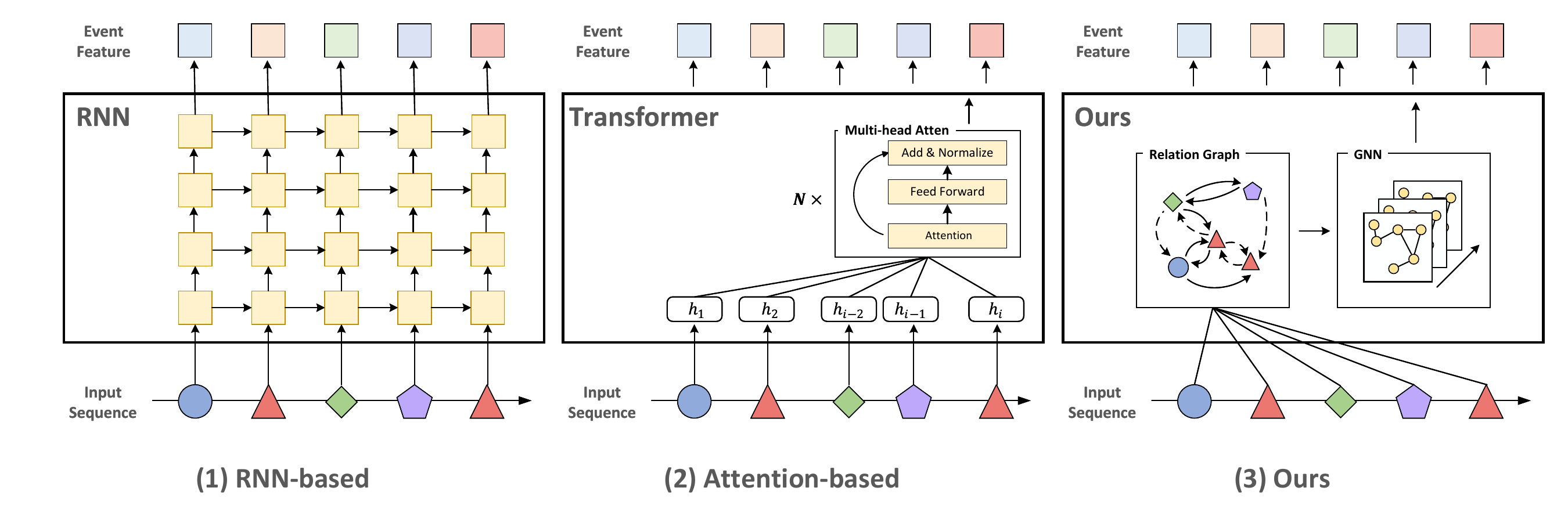}\par\caption{In the asynchronous time series forecasting model, different types of neural temporal point processes (including RNN-based model, Attention-based model and our proposed model) model the correlation between events in the historical sequence .}
\label{fig:fig-1}
\end{figure}

To address the problems above, we propose the Contrastive Relational Inference-based Hawkes Process (CRIHP), utilizing Neural Relational Inference to dynamically model the mutual interactions between events. We formulate the relational inference as a latent variable model, which is learned by variational inference. The latent variables describe the type and strength of interactions between the events \cite{chu2021graph}. Due to the flexibility of latent variable models, a relation graph can be generated as a multi-view graph, which can represent different kinds of interactions, and the generation process is dynamic. To ensure the reliability of the inferred relation graph, we design Contrastive Relational Inference architecture(CRI), which constructs contrasting relationship constraints in the latent space, ensuring that event sequences with similar dynamic patterns also possess similar relation structures \cite{chu2023leveraging,wang2023enhancing}. Furthermore, to accurately identify intent signals during sampling for CRI, we employ intensity-based learning to recognize prototypical paths, which represent the dynamic patterns of event sequence. Our method builds a bridge between the fields of TPP and NRI. A comparison of our proposed model with existing models is shown in Figure~\ref{fig:fig-1}. Extensive experiments on three real-world datasets demonstrate the effectiveness of our proposed model.

%% file: chapters_crihp/_relatedwork.tex
\section{Background}
\label{sec:related_work}

\begin{figure*}[htp]
\centering
\includegraphics[width=0.9\textwidth]{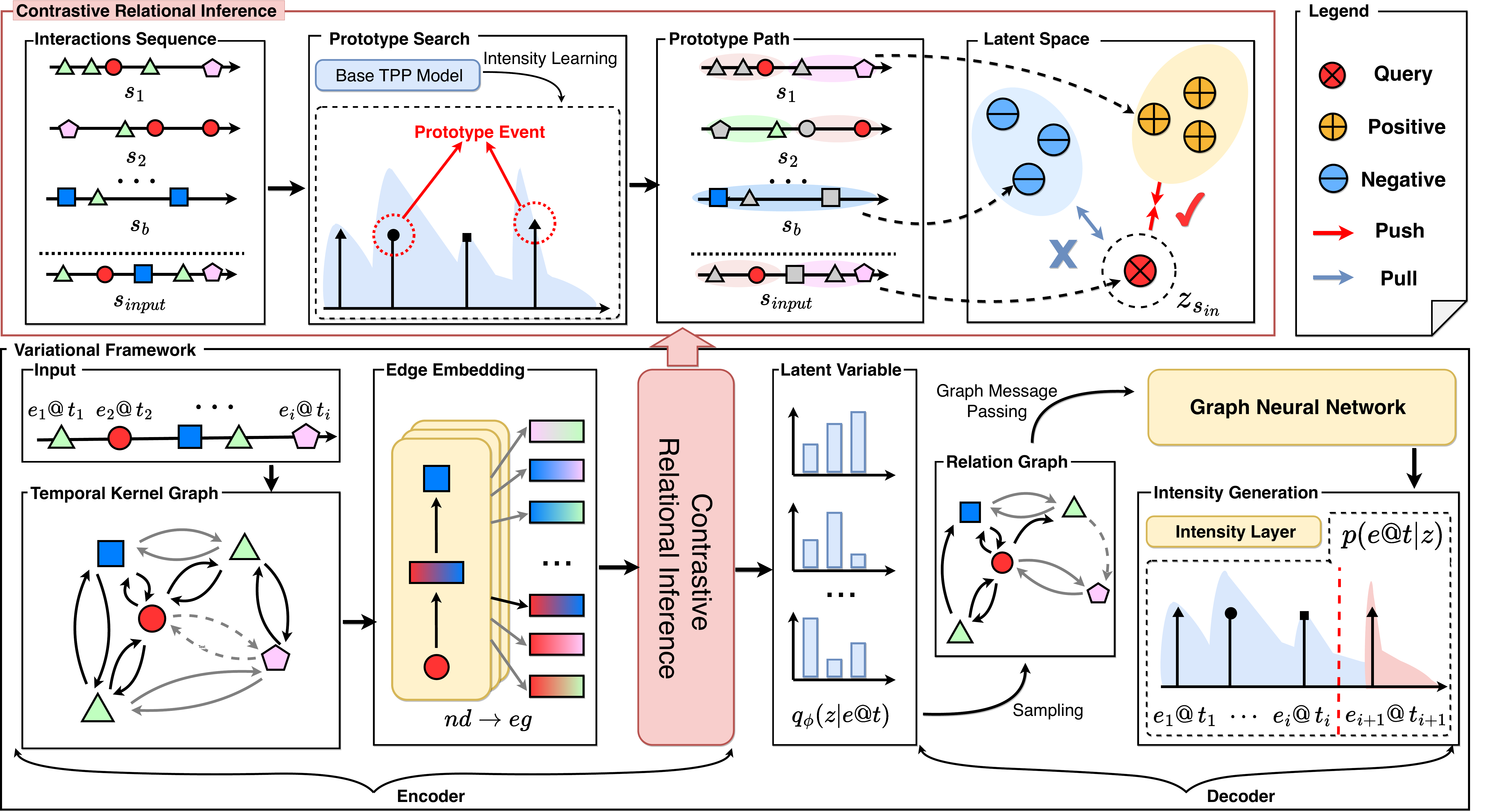}\par
\vspace{-2mm}
\caption{Our framework for asynchronous time series forecasting, Contrastive Relational Inference-based Hawkes Process (CRIHP).}
\vspace{-2mm}

\label{fig:framework}
\end{figure*}

\subsection{Generative Modeling of Event Sequences}
Following the notations in \cite{48:conf/iclr/MeiYE22,51:qu-2022-rltpp}, given a time interval $[0, T]$, we assume $n$ events are observed, composing an asynchronous time event sequence  $s_{[0:T]} = [e_1@t_1, \cdots, e_n@t_n]$ within the time interval. Each event is denoted mnemonically $e_i@t_i$, where, $e_i  \in \{1, ....E \}$ is the discrete event type, and $t_i$ represents the timestamp at which the event occurs, satisfying $0 < t_1 < \cdots < t_n < T$. TPPs model the probability of the next event occurrence by defining a conditional intensity function $\lambda (t)$ and build $\lambda _e$ for each event type $e$, with the objective function defined as follows: 
\begin{equation}
L_{ll} = \sum_{i=1}^{N} log \lambda _{e_i} (t_i | s_{[0,t_i)}) -  \int_{t=0}^{T} \sum_{e=1}^{E} \lambda _e (t | s_{[0,t)]})dt,
\end{equation}
the first term can be regarded as the log-likelihood of observed events, while the second term represents the log-likelihood of non-events. Neural TPP does not involve pre-defined parameterized conditional intensity functions, instead, it utilizes neural networks to learn.

\subsection{Neural Point Processes with GNN}
GNNs provide a direct way to model the dynamics of event sequences. Since the edges in the graph can represent the dependencies between nodes during message passing, GNN-based models naturally realize the relational inference in the event sequence. Existing GNN-based TPPs are mainly divided into two types based on the different node types: including relational graph models based on marks \cite{23:conf/ijcnn/XueSHMZWW21, 27:conf/www/ZhangLY21} and complete events \cite{21:wu2020modeling, 24:conf/kdd/LiLKP21, 26:journals/corr/abs-1909-10367}. The first type of method only constructs relation connections between marks and assumes that events with the same mark information in different historical sequences have the same relationships. This type of method cannot handle the problem of dynamic changes in relationships over time. The second type of method builds relationships based on complete event information. Due to the consideration of dynamic changes in influence relationships, this type of method can construct more realistic relation structures. Similar to the second type of method, our method proposes a generative model for dynamic inference.

%% file: chapters_crihp/_method.tex
\section{Methodology}
\label{sec:method}

In this section, we present the proposed Contrastive Relational Inference-based Hawkes Process (CRIHP) framework. As illustrated in Figure \ref{fig:framework}, similar to variational inference models, CRIHP consists of an encoder and decoder. The encoder performs contrastive relational inference on the input event sequence to generate a relation graph capturing the structure of the events. This inferred relation graph is then utilized by the decoder to forecast future events in a hierarchical manner.

\subsection{CRIHP Framework}
\label{ssec:3-1}
TPPs aim to model the mechanisms that give rise to the dynamics of the recurrence of events. Due to the inherent difficulty in directly observing the interactions between events, in this paper, we dynamically model the interactions within the framework of variational inference. The proposed CRIHP models the relation graph between different events in the historical sequence in the encoder, providing reliable event structures for the decoder. Since the complexity of the relationships between events, it is difficult to achieve effective inference by direct inference. Therefore, we propose a CRI architecture that constrains the generated relation graph in the latent space. Following the curriculum learning, we construct a simpler Front Graph before performing inference,  which allows the model to reason the relation graph in increasing complexity. Specifically, we introduce a temporal similarity graph \cite{24:conf/kdd/LiLKP21} as the Front Graph. For any two events $e_i@t_i$ and $e_j@t_j$ in the historical sequence $s_{[0,T)}$, we calculate the interval time between them and encode it with a kernel function.  We employ the Gaussian kernel function for construction :

\begin{equation}
K\left(t_{i}, t_{j}\right)=\exp (-\frac{\left\|t_{i}-t_{j}\right\|^{2}}{2 \sigma^{2}} ).
\end{equation}

On the basis of a front graph $\Phi^{fro}$, we utilize Graph Convolutional Network to learn the feature embedding of events $H^{fro}$ by $C$, which is the original feature of event sequence obtained by the embedding layer. Then, CRIHP uses the $H^{fro}$ and $\Phi^{fro}$ to further infer the relation graph. The inference process consists of two steps: edge embedding and contrasting relationship inference. In the first step, we merge the embeddings of two adjacent nodes in the front graph and then encode for the connecting edges: 

\begin{equation}
\begin{aligned}
\text{HNd}_{i} &=\operatorname{Rea}_{\text{eg} \rightarrow \text{nd}} (\sum_{i \neq j} \text{HEg}_{(i, j)} ), \\
\text{HEg}_{(i, j)} &=\operatorname{Rea}_{nd \rightarrow eg}\left(\left[\text{HNd}_{i}, \text{HNd}_{j}\right]\right),
\end{aligned}
\end{equation}
$\text{HNd}_i$ represents the embedding of the i-th event, $\text{HEg}_{(i,j)}$ represents the embedding of the connected edge between the i-th and the j-th events. We perform two rounds of node-to-edge and edge-to-node passing processes to aggregate information from multi-hop nodes. Based on this, we obtain the inference results:

\begin{equation}
q_{\phi}\left(\Phi^{rel} \mid \Phi^{fro} \right)= \operatorname{softmax} (CRI \left( \text{HEg})\right),
\end{equation}
where $\Phi^{rel} $ is the relation graph.

In the decoding stage, CRIHP utilizes $\Phi^{rel}$ and $C$ to realize the message passing on the relation graph through GNN and make predictions. To demonstrate the effectiveness of the inferred relation graph,  we employ a two-layer GCN \cite{39:welling2016semi}. During the prediction stage, the Intensity Layer generates the conditional probability distribution for future events.

\subsection{Contrastive Relational Inference}
\label{ssec:3-2}

In the encoding stage,  CRI architecture infers the relation graph $\Phi^{rel}$ based on the front graph. To ensure the reliability of the inferred relation graph, we make the relational consistency assumption. Specifically, we posit that event sequences with similar dynamic patterns exhibit similarity not only in the core events that reflect these patterns but also in the relationship distributions among them. Building upon this assumption, we construct a contrastive learning paradigm in the latent space. As the original event sequences reflect implicit and noisy intention signals, they may not sufficiently capture the actual dynamic patterns of the system. Therefore, the CRI leverages intensity-based learning to search for prototype events in the original sequence and describe the dynamic patterns of the event sequence using the prototype path. The CRI consists of two steps: prototype search and contrasting relationship constraints.

In the prototype search stage, we train an intensity-based TPP as the prototype model to generate the intensity distribution of observed events in sequence. Since the intensity function reflects the probability of an event to a certain extent, we select historical events with higher intensity scores as prototype events $\hat{e_i @ t_i}$ by intensity threshold $\gamma_{I}$, and use these prototype events to construct the prototype path $ PT = \{\hat{e_1 @ t_1}, \ldots, \hat{e_{n_{PT}} @ t_{n_{PT}}} \}$. Based on the prototype path for describing dynamic patterns, we introduce the Optimal Transport Distance (OTD) \cite{46:conf/icml/MeiQE19} to measure the similarity of dynamic patterns between different event sequences, denoted as $d^{otd}_{i,j} = OTD(PT_i, PT_j)$. We sample positive samples for the input sequence $s_{input}$ by $d^{otd}$.

After sampling, we apply contrastive relationship constraints in the latent space. The positive samples $s^{+}$ and negative samples $s^{-}$ are encoded to obtain relation graphs and event embedding, denoted as $z_i = \{\Phi^{rel}_i, C_i\}$. We introduce the normalized temperature-scaled cross-entropy loss (NT-Xent) \cite{47:journals/corr/abs-1807-03748} as the contrastive loss function :

\begin{equation}
\mathcal{L}_{CRI} = -log \frac{exp(sim(z_{i},z^{+})/ \tau)}{\sum^{K}_{j=0}exp(sim(z_{i},z_j^{-})/ \tau)},
\end{equation}
$sim(\cdot)$ represents the cosine similarity function:
\begin{equation}
Sim(z_i, z_j) = z^{\top}_i z_j / || z_i || || z_j ||,
\end{equation}
and $\tau$ denotes the temperature parameter. The contrastive loss is computed in the minibatch.

\subsection{Model Learning}
\label{ssec:3-3}

Since our proposed CRIHP model is based on the framework of variational inference, its objective function follows the evidence lower bound:

\begin{equation}
\begin{aligned}
\mathcal{L} =\mathcal{L}_{ll}+\operatorname{KL}\left[q_{\phi}\left(\Phi^{rel} \mid \Phi^{fro} \right)|| p_{\theta}\left(\Phi^{rel}\right)\right] + \mathcal{L}_{CRI},
\end{aligned}
\end{equation}
which consists of three parts. The first part is the reconstruction error, defined as the log-likelihood of the TPP. The second part is the KL divergence, which can be regarded as a regularization for the base posterior distribution, and we consider a uniform distribution \cite{zhou2023ptse}. If the discrete distribution is sampled, the derivatives can not be backpropagated, so we use the Gumbel Reparametrization \cite{41:conf/iclr/JangGP17} to train the model normally.

%% file: chapters_crihp/_experiment.tex
\section{Experiments}
\label{sec:experiments}

\subsection{Experimental Setup}
\label{ssec:4-1}

\begin{table*}[htbp]
\centering
\vspace{-2mm}
\caption{Performance comparison with neural TPP baselines across three datasets. Higher accuracy (ACC) and lower root mean squared error (RMSE) indicate better model performance.}
\vspace{-2mm}
\label{table:table-1}
\scalebox{1.2}{
\begin{tabular}{ccccccccccc}
\toprule
\multicolumn{2}{c}{Models}   & \multicolumn{2}{c}{RNN-based} & \multicolumn{3}{c}{Attention-based} & \multicolumn{3}{c}{GNN-based} \\
\cmidrule(lr){3-4}\cmidrule(lr){5-7}\cmidrule(lr){8-10}
\multicolumn{2}{c}{Datasets}                                             & RMTPP            & NHP            & THP            & SAHP   & Att-NHP        & GeoHP           & GCHP-GCN          & CRIHP        \\ \midrule
\multicolumn{1}{c|}{\multirow{2}{*}{IPTV}} & \multicolumn{1}{c|}{ACC}  & 56.67 & 50.06  & 72.10          & 71.83  & 73.12       & 43.05          & 75.35        & \textbf{76.72} \\
\multicolumn{1}{c|}{}                       & \multicolumn{1}{c|}{RMSE}  & 22.574 & 18.812 & 12.780          & 13.211   & 11.256     & 20.087         & 10.866       & \textbf{10.131}  \\ \midrule
\multicolumn{1}{c|}{\multirow{2}{*}{ATM}} & \multicolumn{1}{c|}{ACC}  & 76.70 & 73.67 & 70.71          & 67.20   & 75.92      & 21.62        & 90.88      & \textbf{91.95}         \\
\multicolumn{1}{c|}{}                       & \multicolumn{1}{c|}{RMSE}  & 6.221 & 7.031 & 3.820         & 4.591    & 4.130     & 9.014         & 2.612        & \textbf{2.598}      \\ \midrule
\multicolumn{1}{c|}{\multirow{2}{*}{Weeplace}}   & \multicolumn{1}{c|}{ACC}  & 21.97 & 25.17  & 29.10         & 28.65   & 29.38     & 19.22         & 31.81        & \textbf{32.06}     \\
\multicolumn{1}{c|}{}                       & \multicolumn{1}{c|}{RMSE}  & 7.320 & 6.719 & 6.695          & 6.889   & 6.775      & 25.330         & 6.493       & \textbf{6.452} \\ \bottomrule
\end{tabular}
}
\end{table*}

We validate the performance of our CRIHP on multiple real-world asynchronous time series forecasting datasets, including ATM\cite{42:conf/aaai/XiaoYYZC17}, IPTV\cite{43:journals/tbc/LuoXZDXYZ14}, and Weeplace\cite{44:cheng2011toward}. The ATM dataset was provided by 1554 ATMs at a bank in North America, and their event logs of error reporting were recorded. The IPTV data set is provided by China Telecom, which records the sequence of users' viewing behaviors on the network TV. The log information includes the start and end timestamps of each viewing record, the names of the TV programs, and the corresponding category. Weeplace is a Point of Interest (POI) dataset published by Twitter users, and each POI data contains information on geographical location, including latitude and longitude, and area category labels. We extensively compare the CRIHP model with seven existing neural point process models in three categories, including two RNN-based TPP models (including RMTPP \cite{11:conf/kdd/DuDTUGS16}, NHP \cite{12:conf/nips/MeiE17}), three Attention-based TPP models (including THP \cite{15:conf/icml/ZuoJLZZ20}, SAHP \cite{14:zhang2020self}, Att-NHP \cite{48:conf/iclr/MeiYE22}), and two GNN-based TPP models (including GeoHP \cite{45:conf/aaai/ShangS19} and GCHP-GCN \cite{24:conf/kdd/LiLKP21}). We use ACC and RMSE to evaluate the predictive performance of the model for mark information and occurrence time of the next event, and train 200 epochs for each experiment.

\begin{figure}[htp]
\centering
\includegraphics[width=0.45\textwidth]{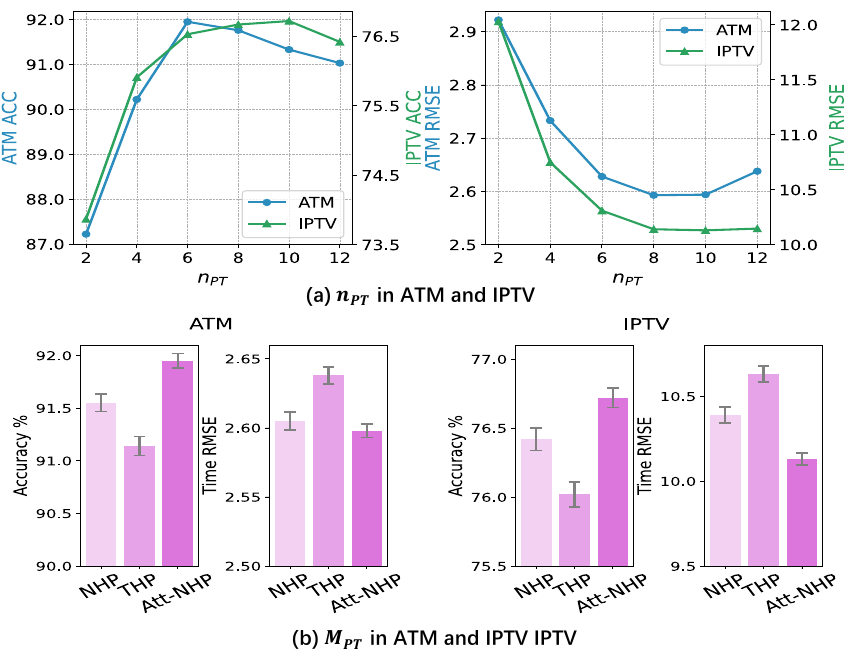}\par
\vspace{-2mm}
\caption{Sensitivity analysis of $n_{PT}$ and $M_{PT}$ on ATM and IPVE.}
\vspace{-2mm}
\label{fig:fig-exp}
\end{figure}

\subsection{Main Results}
\label{ssec:4-2}

Table~\ref{table:table-1} compares the performance of CRIHP against seven baseline models on different datasets. The results show that CRIHP outperforms the other models in predicting the next event marks and times across all three datasets. This advantage stems from CRIHP's effective modeling of interactions between events. The experiments demonstrate inference of event relations enables superior predictive performance compared to RNN and attention-based models lacking relational reasoning. By incorporating variational inference and contrastive learning, CRIHP exhibits greater expressiveness than existing GNN models. Moreover, the advantages of CRIHP over baselines are most pronounced on the IPTV dataset. The richer event marks in IPTV increase the difficulty of forecasting, which CRIHP handles well through its relational approach to event structure modeling. Overall, CRIHP's relational modeling of complex inter-event effects underlies its strong performance on event sequence prediction tasks.

\begin{table}[htbp]
 \centering
 \vspace{-2mm}
 \caption{Ablation study of the latent variable model, front graph, and CRI.}
 \vspace{-2mm}
 \label{table:table-2}
 \scalebox{1.2}{
 \begin{tabular}{ccccc} 
   \toprule 
    Dataset & \multicolumn{2}{c}{ATM} & \multicolumn{2}{c}{IPTV} \\
    \cmidrule(lr){2-3}\cmidrule(lr){4-5}
    Model & ACC & RMSE & ACC & RMSE\\
   \midrule w/o LVM & 90.60 & 2.636 & 75.21 & 10.878 \\ 
            w/o  front graph & 89.33 & 3.292 & 74.73 & 11.342 \\ 
   \midrule w/o CRI & 89.56 & 3.174 & 74.65 & 11.002 \\
            w/o prototype search & 90.72  & 2.832 & 75.40 & 10.432\\ 
   \midrule Ours & \textbf{91.95} & \textbf{2.598} & \textbf{76.72} & \textbf{10.131} \\
   \bottomrule 
 \end{tabular} }
\end{table}


\subsection{Ablation Study}
\label{ssec:4-3}
To demonstrate the effectiveness of our proposed CRIHP, we conduct ablation studies using two benchmark datasets: ATM and Weeplace. First, we evaluated the effectiveness of the latent variable model and the front graph. Based on the CRIHP model, we remove the NRI and temporal kernel graph respectively, As shown in Table~\ref{table:table-2}, our proposed relational Inference method makes the model infer inter-event relationships more effectively and has better predictive performance. We also verified the effectiveness of the proposed CRI and completely removed the CRI. Additionally, to demonstrate the effectiveness of the sampling method based on the prototype path, we retained the contrastive relationship constraints but removed the prototype search. Instead, we directly constructed the OTD distances between original event sequences for contrastive learning sampling. The experimental results demonstrate that, compared to not using this structure, CRI effectively enhances the reliability of relational inference.

We evaluated the sensitivity of CRIHP to two key parameters: the length of prototype path $n_{PT}$ and the type of the prototype model $M_{PT}$. Regarding $n_{PT}$, as shown in Figure ~\ref{fig:fig-exp}, on the ATM, the model's performance starts to decrease from $n_{PT} = 6 $; on the IPTV, the performance starts to decrease from $n_{PT} = 10$. This indicates that as the $n_{PT}$ increases, the prototype path becomes more similar to the original sequence. Additionally, the inflection point of CRIHP on the ATM dataset occurs earlier than on IPTV, which is related to the shorter average sequence length in ATM. We also analyzed the sensitivity of CRIHP to the choice of prototype model. We selected NHP, Att-NHP, and THP as the base TPP model. As shown in Figure ~\ref{fig:fig-exp}, when selecting Att-NHP, CRIHP exhibits the best predictive performance.